\documentclass[conference]{IEEEtran}

\usepackage{algorithmic}
\usepackage[linesnumbered,ruled,vlined]{algorithm2e}
\usepackage{array}
 \usepackage{verbatim}
\usepackage{amsthm}
\usepackage{amssymb}
\usepackage{amsmath}
\usepackage{graphicx}
\usepackage{listings}
\usepackage{tikz}
\usepackage{color}
\usepackage{url}
\usepackage{zeta}

\definecolor{olivegreen}{rgb}{0.2,0.8,0.5}
\definecolor{grey}{rgb}{0.5,0.5,0.5}

\newtheorem{example}{Example}

\lstdefinelanguage{ttl}{
sensitive=true,
showspaces=false,
showstringspaces=false,
keywords={author, birthPlace, birthDate, broader,subject, label, completionDate, type, genre, philosophicalSchool,publicationDate, mainInterest, movement},
morecomment=[l][\color{grey}]{--},
morestring=[b][\color{blue}]\",
morecomment=[s][\color{olivegreen}]{<}{>},
alsoletter={-,<,>},
emph={dbo,dbr,dc, Category, rdf, rdfs, skos}, emphstyle=\itshape,
emph={[2]fun,cat, lincat,lin},emphstyle={[2]\color{red}}
}

\ifCLASSOPTIONcompsoc
  \usepackage[caption=false,font=normalsize,labelfont=sf,textfont=sf]{subfig}
\else
  \usepackage[caption=false,font=footnotesize]{subfig}
\fi

\begin{document}
\title{Mining Arguments from Cancer Documents Using Natural Language Processing and Ontologies}

\author{\IEEEauthorblockN{Adrian Groza\IEEEauthorrefmark{1}, Popa Oana Maria\IEEEauthorrefmark{1},
\IEEEauthorblockA{\IEEEauthorrefmark{1}Intelligent Systems Group, Department of Computer Science,\\
Technical University of Cluj-Napoca, Romania,\\
Adrian.Groza@cs.utcluj.ro, Oana.Popa@cs-gw.utcluj.ro}
}
}

%\author{}

\maketitle

% As a general rule, do not put math, special symbols or citations
% in the abstract
\begin{abstract}
In the medical domain, the continuous stream of 
scientific research contains contradictory results supported by arguments and counter-arguments.
As medical expertise occurs at different levels, part of the human agents have difficulties to face the huge amount of studies, 
but also to understand the reasons and pieces of evidences claimed by the proponents and the opponents of the debated topic.
To better understand the supporting arguments for new findings related to current state of the art in the medical domain we need tools
able to identify arguments in scientific papers. 
Our work here aims to fill the above technological gap.

Quite aware of the difficulty of this task, we embark to this road by relying on the well-known interleaving of domain knowledge with natural language processing. 
To formalise the existing medical knowledge, we rely on ontologies.
To structure the argumentation model we use also the expressivity and reasoning capabilities of Description Logics.
To perform argumentation mining we formalise various linguistic patterns in a rule-based language.
We tested our solution against a corpus of scientific papers related to breast cancer. 
The run experiments show a F-measure between 0.71 and 0.86 for identifying conclusions of an argument and between 0.65 and 0.86 for identifying premises of an argument. 
\end{abstract}

% no keywords


\section{Introduction}
\label{sec:intro}
%Healthcare industry today generates large amounts of complex data about patients, hospitals resources, disease diagnosis, electronic patient records, medical devices etc. 

Consider the recent contradictory results on cancer published in the distinguished journals Science and Nature.
On the one hand we have the advocaters of the so called ``back-luck of cancer''. 
The study in~\cite{tomasetti2015variation} supports the idea that random mutations in healthy cells may explain two-thirds of cancers. 
These results suggest that most cancer cases can not be prevented. 
One positive side of this randomness of cancer is that it helps cancer patients to know that is not their fault~\cite{couzin2015bad}.
The intriguing correlations discovered by Tomasetti and Vogelstein contradict the older landmark paper~\cite{doll1981causes} of Doll and Peto arguing that 
most cancers could be prevented by removing various lifestyles. 
On the other hand, the advocaters of risk factors of cancer provide set of counter-arguments against~\cite{tomasetti2015variation} through the voices of Wodarz and Zauber~\cite{wodarz2015cancer}. 

The above example is a good instantiation of the problems arising by continuous stream of 
scientific research that contains contradictory results supported by arguments and counter-arguments.
As medical expertise occurs at different levels, part of the human agents have difficulties to face the huge amount of studies, 
but also to understand the reasons and pieces of evidences claimed by the proponents and the opponents of debate topic.
To better understand the supporting arguments for new findings related to current state of the art in the medical domain we need tools
able to identify arguments in scientific papers. 
Our work here aims to fill the above technological gap.

Quite aware of the difficulty of this task, we embark to this road by relying on the well-known interleaving of domain knowledge with natural language processing. 
To formalise the existing medical knowledge, we rely on ontologies.
To structure the argumentation model we use also the expressivity and reasoning capabilities of Description Logics.
To perform argumentation mining we formalise various linguistic patterns in a rule-based language.
We tested our solution against a corpus of scientific papers related to breast cancer. 

In the breast cancer domain, where monthly appear more and more articles, this being a disease very 
spread across women in many countries. 
The recent proliferation of the on-line publication of medical research articles 
has created a critical need for information access tools that help stakeholders in the medical domain.

%\textit{Current limitation of mining medical documents} : 
Because of the amount of information about a particular subject, data mining brings a set of tools and techniques that can be applied to this processed data to discover hidden patterns. 
This provides healthcare professionals an additional source of knowledge for making decisions. 
Current limitations or challenges in data mining for healthcare include information from heterogeneous sources present challenges or missing values, noise and outliers.

%\textit{Our research hypothesis}: 

%Follow this pattern: To learn X we did Y
We proposed argumentation as the underlying technological instrumentation having the purpose of helping healthcare professionals for supporting decision making. 
This research focus on understanding by generating cognitive maps or argumentation graphs. % with the purpose of construction and accumulation of spatial knowledge.
% allowing the "mind's eye" to visualize images in order to reduce cognitive load, enhance recall and %learning of information.
Argumentation is the process where arguments are structured and evaluated based on the their interactions with each other~\cite{Letia2006,Letia2012}. An argument consists of a set of premises, offered with the purpose of supporting the claim. %The claim is a proposition that is always taken as true, even if is true or false.
Argumentation may also involve chains of reasoning, where claims are used as premises for deriving further claims. %Argumentation plays an important role in many areas, but especially in the artificial intelligence and Natural Language Processing (NPL).
Argumentation mining~\cite{mochales2011argumentation,peldszus2013argument} is a new research area that combines Natural Language Processing (NPL) with the argumentation theories and question answering. 
Argumentation mining aims to automatically detect arguments from text documents, including other functionalities like the structure of an argument and the relationship between them.

The remaining of this paper is structured as follows: 
Section~\ref{sec:technical} introduces the technical instrumentation used throughout the paper.
Section~\ref{sec:architecture} details the architecture of the system.
%Section~\ref{sec:solution} describes the proposed solution, 
Section~\ref{sec:experiment} details the running experiments. 
Section~\ref{sec:related} discusses related work and section~\ref{sec:con} concludes the paper.

\section{Argumentation model}
\label{sec:technical}
This section formalises in Description Logic (DL) the argumentation model.
First, we introduce the basic terminology of DLs.
Second, we detail the argumentation model used for the argumentation mining task.

\subsection{Description logics}
\label{subsec:descriptionLogic}
In Description Logics (DLs) concepts are built using the set of constructors formed by
negation, conjunction, disjunction, value restriction, and existential restriction~\cite{baader2003description} 
(Table~\ref{tab:dl}). 
Here, $C$ and $D$ represent concepts and $r$ is a role.
The semantics is defined based on an interpretation $I = (\Delta^{I}, \cdotp^{I})$, where the domain $\Delta^{I}$ of $I$ contains a non-empty set of individuals, and the interpretation function $\mathit{\cdotp^I}$ maps each concept $C$ to a set of individuals $C^I\in \Delta^I$ and each role $r$ to a binary relation $r^I\in \Delta^I \times \Delta^I$. 
The last column of Table~\ref{tab:dl} shows the extension of $\cdotp^{I}$ for non-atomic concepts.

\begin{table}
\begin{footnotesize}
\caption{Syntax and semantics of $\mathcal{ALC}$.}
\label{tab:dl}
\begin{tabular}{|l|l|l|}
\hline
{\it Constructor} & {\it Syntax} & {\it Semantics} \\
\hline
%negation & $\mathit{\neg C}$ & $\mathit{\Delta^{I} \setminus C^I}$\\ \hline 
negation & $\neg C$ & $\Delta^{I} \setminus C^I$\\ \hline 
conjunction & $C \sqcap D$ & $C^I \cap D^I$\\ \hline
disjunction & $C \sqcup D$ & $C^I \cup D^I$\\ \hline
existential restriction & $\exists r.C$ & $\{x\in \Delta^I | \exists y: (x,y)\in r^I \wedge y\in C^I\}$ \\ \hline
value restriction & $\forall r.C$ & $\{x\in \Delta^I | \forall y: (x,y)\in r^I \rightarrow y\in C^I \}$ \\ \hline 
individual assertion & $a:C$ & $\{a\} \in C^I$\\ \hline
role assertion & $(a,b):r$ & $(a,b)\in r^I$\\ \hline
\end{tabular}
\end{footnotesize}
\end{table}

%\begin{definition}
A knowledge base (KB) is formed by a terminological box $TBox$ and an assertional box $ABox$.
The $\mathit{TBox}$ contains terminological axioms of the forms $C \equiv D$ or $C \sqsubseteq D$.
The $\mathit{ABox}$ represents a finite set of concept assertions {\it a:C} or role assertions {\it (a,b):r}, where $C$ is a concept, $r$ a role, and $a$ and $b$ are two individuals.
A concept $C$ is satisfied if there exists an interpretation $I$ such that $C^I \neq \emptyset$.  
%The concept $D$ subsumes the concept $C$ ($\mathit{C\sqsubseteq D}$) if $\mathit{C^I \subseteq D^I}$ for all interpretations $I$.
The concept $D$ subsumes the concept $C$, represented by $C \sqsubseteq D$ if $\mathit{C^I \subseteq D^I}$ for all interpretations $I$.
%\end{definition}
Constraints on concepts (i.e. {\it disjoint}) or on roles 
({\it domain}, {\it range}, {\it inverse} role, or {\it transitive} properties) can be specified in more expressive description logics\footnote{We provide only some basic terminologies of description logics in this paper to make it self-contained. For a detailed explanation about families of description logics, the reader is referred to~\cite{baader2003description}.}.

\begin{figure}
\small
\begin{align}
ClinicalArgument \sqsubseteq Argument\\
\exists supports . \top \sqsubseteq Argument\\
\top \sqsubseteq \forall supports . Argument\\
supports^+\\
\exists attacks . \top \sqsubseteq Argument\\
\top \sqsubseteq \forall attacks . Argument
\end{align}
\caption{TBox example in the argumentation domain.}
\label{fig:tbox}
\end{figure}

The Tbox in Fig.~\ref{fig:tbox} introduces the subconcept \textit{ClinicalArgument} which is a particular type of argument.
An argument can support another argument via the role \textit{supports} that has as domain the concept \textit{Argument} (line 2) and the same range \textit{Argument} in line 3.
Line 4 specifies that the role \textit{supports} is transitive.
The attack relationship between two arguments is modeled by the role \textit{attacks} which has as domain and range the set of arguments (lines 5 and 6).
The following Abox contains the individual $a$ of type \textit{ClinicalArgument} which has the premise $p$ and the claim $c$.
%\begin{figure}
$a:ClinicalArgument$, $(a,p):hasPremise$, $(a,c):hasClaim$.
%1   \hspace*{5mm}(implies MedicalArgument argument)\\
%2   \hspace*{5mm}(implies ClinicalArgument MedicalArgument)\\
%3   \hspace*{5mm}(instance a ClinicalArgument)\\
%4   \hspace*{5mm}(related a p1 hasPremise)\\
%5   \hspace*{5mm}(related a p2 hasPremise)\\
%6   \hspace*{5mm}(related a c1 hasClaim)\\
%7   \hspace*{5mm}(concept-instances (at-least 1 hasPremise))\\
%8   \hspace*{5mm}(concept-instances (at-least 1 hasClaim))\\
%9   \hspace*{5mm}(concept-instances (exactly 1 hasClaim))
%}
%\caption{Abox with a clinical argument.}
%\label{fig:abox}
%\end{figure}

\subsection{Breast cancer ontology}

We are interested in breast cancer ontologies.
%In this paper, the focus is on medical ontologies and more specific to cancer ontologies, breast cancer ontologies. 
The \textit{Breast Cancer Grading Ontology} (BCGO) assigns a grade to a tumor starting from the three criteria of the NGS, being part of the Biological Process category.
%Ontology provides a common framework for structured knowledge representation of domain knowledge. 
%Ontology framework provides common vocabulary for concepts, in this case medical concepts, concept definitions, relationships, axioms and rules, allowing a controlled flow of knowledge into the knowledge base.

The Tbox in Fig.~\ref{fig:tboxOfCancer} introduces concepts like \textit{Cancer} which is a particular type of \textit{Disease} and \textit{BreastCancer} is a particular type of \textit{Cancer} (axioms~\ref{eq:1}, \ref{eq:2}).
A disease has symptoms, presented by the role \textit{manifestedSymptom} that has as range the concept \textit{Symptom} (axiom~\ref{eq:3}). 
One or more treatments can be recommended  via the role \textit{appliedTreatment} that has as range the concept \textit{Treatment} (axiom~\ref{eq:4}). 
Breast cancer heavily affects all fields of the human life. 
This is modeled by the role \textit{affectedDomain} \textit{Domain} (axiom~\ref{eq:5}).
 
People are implied here, like doctors and patients, this being presented by the \textit{impliedPerson} role, which has as range the concept \textit{Person} (axiom \ref{eq:6}). Breast cancer has characteristics, this being modeled by the role \textit{haveCharacteristic} with range in the concept \textit{Characteristic} (axiom \ref{eq:7}). In some cases the people involved of affected by this disease are numbered and for this is used the role \textit{haveQuantifier} with the domain \textit{People} and the range \textit{Quantifier} (axioms \ref{eq:8}, \ref{eq:9}).

%\begin{example}[TBox example]
\begin{figure}
\small
\begin{align}
Cancer \sqsubseteq Disease \label{eq:1}\\ 
BreastCancer \sqsubseteq Cancer \label{eq:2}\\ 
\top \sqsubseteq \forall manifestedSymptom . Symptom \label{eq:3}\\
\top \sqsubseteq \forall appliedTreatment . Treatment \label{eq:4}\\
\top \sqsubseteq \forall affectedDomain . Domain \label{eq:5}\\
\top \sqsubseteq \forall impliedPerson . Person \label{eq:6}\\
\top \sqsubseteq \forall haveCharacteristic . Characteristic \label{eq:7}\\
\exists haveQuantifier . \top \sqsubseteq People \label{eq:8}\\
\top \sqsubseteq \forall haveQuantifier . Quantifier \label{eq:9}
\end{align}
\caption{TBox example in the cancer domain.}
\label{fig:tboxOfCancer}
\end{figure}

The Abox in Fig.~\ref{fig:aboxCancer} contains the individual $a$ of type \textit{BreastCancer} instantiated with \text{\textit{"Angiosarcoma"}}. This individual manifest the symptom \text{\textit{"Skin irritation or dimpling"}} and implied the persons \text{\textit{"Doctors"}} and \text{\textit{"Woman"}}. The treatment applied for this instance is the individual \text{\textit{"Chemotherapy"}}. Also, this disease affected the individual domain \text{\textit{"Family history"}} and have as characteristics \text{\textit{"Hormone receptivity"}} and \text{\textit{"High levels of HER2"}}. 

\begin{figure}
\small
\begin{align}
a: BreastCancer\\
(a,\text{ \textit{"Skin irritation or dimpling"}}):manifestedSymptom\\ 
(a,"Chemotherapy"):appliedTreatment\\ 
(a,\text{ \textit{"Family history"}}):affectedDomain\\ 
(a,"Doctors"):impliedPerson\\ 
(a,"Woman"):impliedPerson\\
(a,\text{\textit{"Hormone receptivity"}}):haveCharacteristic\\
(a,\text{\textit{"High levels of HER2"}}):haveCharacteristic
\end{align}
\caption{Abox with an instance of breast cancer disease. Here a is a shortcut of "Angiosarcoma".}
\label{fig:aboxCancer}
\end{figure}
%Ontology reasoners help researchers by mining inferred knowledge from the knowledge represented
%in the ontology.

The system uses cancer ontology to build more specific lists of words. 
The terminology is input to text files such as \textit{cancerRelatedWords.lst} for terms relating to the cancer domain and \textit{peopleInvolved.lst} for terms that may indicate people involved or affected by this disease. 
The lists are used by a gazetteer that associates the terms with a majorType such as \textit{"CancerRelatedWords"} or \textit{"PeopleInvolved"}. 
JAPE rules convert these to annotations that can be visualised and queried. 

For example, suppose a text has a token term \textit{"breast cancer"} and GATE has a gazetteer list with \textit{"breast cancer"} on it; GATE (see~\cite{cunningham2002gate}) finds the string on the list, then annotates the token with majorType as \textit{"CancerRelatedWords"}; we convert this into an annotation that can be visualised or searched such as \textit{CancerRelatedWords}. A range of terms that may indicate people involved are all annotated with \textit{"PeopleInvolved"}. 

%One way of querying the cancer ontology is with the command \textit{(CONCEPT-INSTANCES  $individualName$)} where the $individualName$ is filled with $TREATMENT$, $SYMPTOM$, $CHARACTERISTIC$, $PEOPLE$, $DOMAIN$. Other query that can be used is \textit{(INDIVIDUAL-TOLD-ATTRIBUTE-VALUE $individualName$ $attributeName$)}. This helps the human user to obtain the attribute of an $individualName$, where the $attributeName$ is filled with $HASTEXT$.

The tool can also create annotations for complex concepts out of lower level annotations. 
In this way, the gazetteer provides a cover concept for related terms that can be queried or used by subsequent annotation processes.
%In the implementation, the system has gazetteer lists for cancer domain terminology and for user domain %terminology, one list each for conclusions, premises, and a range of verbs, factors, domains affected %terminology lists. 
The advantage of using ontologies for making the lists specified above is that they can help build more powerful and more interoperable information systems in healthcare. %Ontologies can support the need of the healthcare process to transmit, re-use and share patient data. 

%Ontologies can also provide semantic-based criteria to support different statistical aggregations for %different purposes.

%Possibly the most significant benefit that ontologies may bring to healthcare systems is their ability %to support the indispensable integration of knowledge and data.

%But on the negative side, some remain skeptical about the impact that ontologies may have on the design %and maintenance of real-world healthcare information systems.

\subsection{Argumentation model}

%We define an argument in definition~\ref{def:arg}:
%	\begin{definition}
	An argument $a\langle p_i, c\rangle$ contains an exactly one conclusion $c$ and a set of supporting premises $p_i$. The definition in DL follows: 
	\begin{small}
	\begin{equation}
	 Argument \equiv \exists\ hasPremise.Premise \sqcap (=1)hasClaim.Claim
          \label{eq:arg}
	\end{equation}
	\end{small}
	\label{def:arg} 
%	\end{definition}

	We assume that claims and premises have textual descriptors and are signaled by specific lexical indicators.
\begin{small}
	\begin{align}
Claim  \equiv & \exists hasText.String \sqcap \exists hasIndicator.ClaimIndicator\\
Premise  \equiv & \exists hasText.String \sqcap \exists hasIndicator.PremiseIndicator 
\end{align}
\end{small}

We rely on textual indicators classified in several concepts:
 
\begin{small}
\begin{align}
Indicator \equiv PremiseIndicator \sqcup ClaimIndicator \sqcup MacroIndicator
\end{align}
\end{small}

Inheritance between roles has been enacted for subroles \textit{hasPremiseIndicator} and \textit{hasClaimIndicator}:

\begin{small}
\begin{align}
\top \sqsubseteq \forall hasIndicator . Indicator\\
hasPremiseIndicator \sqsubseteq hasIndicator\\
hasClaimIndicator \sqsubseteq hasIndicator
\end{align}
\end{small}

\begin{example}[Sample of claim indicators]
	The following lexical indicators usually signal a claim and they are instances of the concept \textit{ClaimIndicator} (Ci): "consequently", "therefore", "thus", "so", "hence", "accordingly", "we can conclude that", "it follows that", "we may infer that", "this means that", "it leads us to believe that", "this bears out the point that", "which proves/implies that", "as a result" %"often", "the purpose of", "results of such", "the fact that". 
\label{ex:claim}
	\end{example}

\begin{example}[Sample of premise indicators]
	The following expressions usually signal a premise and they are instances of the concept \textit{PremiseIndicator} (Pi): "since", "because", "for", "whereas", "in as much as", "for the reasons that", "in view of the fact", "as evidenced by", "given that", "seeing that", "as shown by", "assuming that", "in particular". 
	\label{ex:premise}
	\end{example}

The \textit{MacroIndicator} is formed by several words, one of then indicating verbs related to claim or premise: 
%We rely on the following inheritance: 
\begin{small}
\begin{align}
VerbRelatedToClaim \sqcup VerbRelatedToPremise \sqsubseteq MacroIndicator
%VerbRelatedToPremise \sqsubseteq MacroIndicator
\end{align}
\end{small}

\begin{example}[Verbs related to claim]
	The following verbs usually signal a claim and they are instances of the concept \textit{ConclusionVerbs} (Vc): "to report", "to believe", "to assesse", "to identify", "to highlight", to be essential", "to confirm", "to estimate", "to provide", "to express", "to experience", "to recall", "to accept", "to reflect", "to categorize", "to indicate", "to exemplify", "to define", "to show", "to qualify"%, "to increase", "to include", "to describe", "to be equal", %"to speak", "to know".  
\label{ex:VerbClaim}
	\end{example}

	\begin{example}[Verbs related to premise]
	The following verbs usually signal a premise they are instances of the concept \textit{PremiseVerbs} (Vp): "to note", "to subdivide", "to contain", "to result", "to observe", "to accord", "to regard", "to feel", "to show", "to receive", "to examine", "to report", "to transcribe", "to encompass". \label{ex:VerbPremise}
	\end{example}

	Note that the premise indicators might appear after the conclusion was stated, as example~\ref{ex:next} illustrates.

\begin{example}[Premise indicator before the premise] Consider the phrase: 
\begin{small}
\begin{quotation}
"[Spirituality was highlighted as a fundamental component of the healing process]$_{Claim}$. 
[[In particular]$_{PremiseIndicator}$, survivors noted that their faith in God’s direction over the doctors healed them.]$_{Premise}$	
\end{quotation}
\end{small}

The text is annotated with the claim and premise which has a premise indicator (PI), namely \textit{"In particular"}.
The argument has its claim on the first position and one premise following the claim. 
The premise indicator precedes its premise. 
The corresponding Abox follows:

\begin{small}
\begin{align}
(a,c):hasClaim, (a,p):hasPremise\\
(p,"In\ particular"):hasPremiseIndicator\\
(c,p):before, ("In\ particular",p):before
\end{align}
\end{small}
Based on the above identified information, the system classifies argument $a$ as \textit{ClaimPremiseArgument}.
\label{ex:next}
\end{example}

Consider the medical argument in example~\ref{ex:arg} :
\begin{example}[Argument example] $ $

\begin{small}
\begin{quotation}
[Key informants highlighted]$_{ClaimIndicator}$ spirituality as a very important component of many women's cancer experience.
These communities, particularly African American, Asian and Latina, hold firm religious and spiritual beliefs and practices.
[[In particular]$_{PremiseIndicator}$, many have an unshakable belief in the power of prayer, putting more importance on spirituality, their religious beliefs than on health care providers.]$_{Premise}$
\end{quotation}
\end{small}

The argument structure is formalised by:

\begin{small}
\begin{align}
	a:Argument, (a,c):hasClaim, (a,p)hasPremise\\
	(c,\text{"Key informants highlighted")}:hasIndicator\\
	%(p1,"In particular, many have an unshakable\\ belief in the power of prayer, putting more importance on spirituality, their religious beliefs than on health care providers.")&:&hasText\\
	(p, \text{``In particular"}):hasIndicator
\end{align}
\end{small}
\label{ex:arg}
\end{example}
	
There are arguments in which premise precedes the conclusion, 
but also arguments in which the premise appears after the claim.
\begin{small}
\begin{align}
PCArgument \equiv \exists hasPremise.(\exists before.Claim) \\
CPArgument \equiv \exists hasPremise.(\exists after.Claim) 
\end{align}
\end{small}

The roles $before$ and $after$ are inverse and transitive roles, with both the domain and range represented by sentences:

\begin{small}
\begin{align}
before^- \equiv after \\
before^+ \\
after^+\\
\top  \sqsubseteq  \forall before.Sentence\\
\exists before.\top \sqsubseteq Sentence
\end{align}
\end{small}

Instances of $PCArgument$ and $CPArgument$ are illustrated in examples~\ref{ex:before} and~\ref{ex:after}. %, we will give some text where the structure of an argument consist of a claim and premises, but arranged in different order. 

\begin{example}[PCArgument] As premise appears before the claim, the identified argument is classified as a \textit{PCArgument}: 
\begin{small}
\begin{quotation}
"[For women with non proliferative findings, no family history, a weak family history of breast cancer]$_{Premise}$, [doctors reported no increased risk."]$_{Claim}$
\end{quotation}
\end{small}
\label{ex:before}
\end{example}

\begin{example}[CPArgument] Consider the text:
\begin{small}
\begin{quotation}
"[[Patients report]$_{MacroIndicator}$ on the risk of breast cancer]$_{Claim}$
[according to histologic findings, the age at diagnosis of benign breast disease, the strength of the family history."]$_{Premise}$
\end{quotation}
\end{small}
This sentences presents first the $Claim$ part introduced by the macro identifier "Patients report", followed by the $Premise$ part. Hence, this argument is classified as a \textit{CPArgument}.
\label{ex:after}
\end{example}

\section{System architecture}
\label{sec:architecture}

%The system architecture is heavily influenced by the pipeline approach in which Natural Language Processing (NPL) ~\cite{NLP} techniques are to be applied on the input documents.
 
\begin{figure}
\centering
\includegraphics[width=0.5\textwidth]{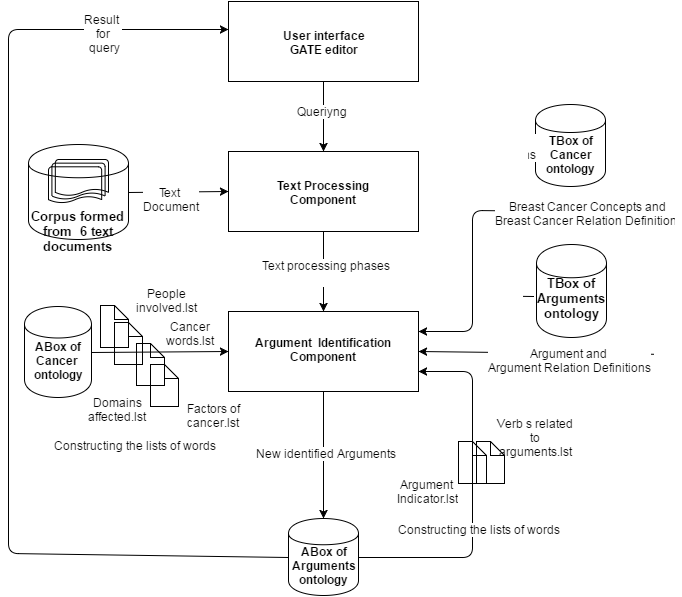}
\caption{Architecture of the system}
\label{fig:system}
\end{figure}

%The architecture of the system is present in Fig. 3. 
The developed argumentation mining system in Fig.~\ref{fig:system} has four components: the Gate editor, text processing component, argument identification modules and the knowledge module. 

% composed out of a TBox and ABox one for the argument ontology and another for the breast cancer ontology. The system works closely with the Text Corpus created formed by six documents.

The first layer consists of the GATE Editor~\cite{cunningham2002gate} and a query interface for the updated ontology. 
The second layer is composed by the text processing component performs the Natural Language Processing transformations required for extracting arguments. %from the input text, example of input ~\cite{exampleArt}.
The argument processing modules aims to identify argumentative sentences in the text. 
Using the TBox and the ABox the system can save in the ontology the structure of the new arguments. 
The cancer ontology is used for creating the lists of words used inside the JAPE rules for identifying the argument structure.
The TBox related to arguments stores the definitions of an \textit{Argument} formed by \textit{Claim} and one or more \textit{Premise}. The ABox related to arguments contains the character instances the application found in the text document. This TBox is used to generate the lists of \textit{ClaimIndicator}, \textit{PremiseIndicator}, \textit{VerbRelatedToClaim} and \textit{VerbRelatedToPremise}.

We apply the tool to the detection of arguments from different articles within the breast cancer domain. 

%GATE~\cite{cunningham2002gate} engineering tool was enacted for natural language processing based on the ANNIE pipeline: tokenizer, POS tagger, gazetteer and JAPE transducer.
%The \textit{Tokeniser} splits the text in simple tokens, such as numbers, punctuation and words of various kinds. The ANNIE Sentence \textit{Splitter} segments text into sentences, while the \textit{Tagger} produces labels corresponding for each part of speech of each word or symbol.
%The lookup list contained by the \textit{Gazetteer} is used in the initial phase of the detection rules of annotations and processing resources. 
%The framework has the ability to cascade Gate JAPE chained rules.

Racer~\cite{HHMW12} was used to perform reasoning in DL and query the system. %RacerPro (Renamed ABox and Concept Expression Reasoner Professional) is a server able to reason on the ontologies loaded on it. RacerPorter is a Graphical User Interface running as a client for RacerPro.
Using Racer, the system saves the newly arguments detected in the breast cancer documents into an Abox. 
%This database is formed by a TBox and an ABox. In the ABox the system can find the properties of an argument, also the relations between them and the attributes are specified. In a TBox are present the instances detected in texts, each of them being unique.
The resulted ontology is used for query answering. %This queries are identified in the interface by buttons with suggestive names. In Eclipse the ontology is loaded into the 

	\subsection{Jape rules}

For identifying the claim and the premises, we use JAPE (Java Annotation Patterns Engine) rules~\cite{cunningham2002gate}. 
%JAPE provides finite state transduction over annotations based on regular expressions.
%JAPE allows to recognise regular expressions in annotations inside documents. 
A JAPE grammar consists of a set of phases, each of which consists of a set of pattern/action rules. %The phases run sequentially and constitute a cascade of finite state transducers over annotations. 
The left-hand-side (LHS) of the rules consists of an annotation pattern description. 
The right-hand-side (RHS) consists of annotation manipulation statements. 
Annotations matched on the LHS of a rule may be referred to on the RHS by means of labels that are attached to pattern elements.

\begin{algorithm}
\label{alg:eval}
\caption{Argumentation mining with patterns.}

\KwIn{$\mathcal{O}$ - breast cancer ontology;
      $\mathcal{A}(Tbox)$ - argumentation model (arg Tbox);\\
      $\mathcal{C}$ - corpus of medical documents}
\KwOut{$\mathcal{A}$, Abox containing mined arguments}

\ForEach{$d \in C$}{
$Sentences \leftarrow Tokenise(d)$

\ForEach{$s\in Sentences$}
{
	if $\exists CC \in s$ then { \\
	\hspace*{2mm}if ClaimMacro $\in s$ then { \\
			\hspace*{4mm}if $ofstClaimMacro \leq  ofstCC$ then { \\
					\hspace*{6mm}$Claim \leftarrow words[ofstClaimMacro, ofstCC]$ \\
					\hspace*{6mm}if PremiseMacro $\in s$ then  { \\
							\hspace*{6mm}$Premise \leftarrow words[ofstCC, ofstPremiseMacro]$ }\\
						} \hspace*{4mm}else {\\ 
						  \hspace*{4mm} $Claim \leftarrow words[ofstCC, ofstClaimMacro]$ \\
						  \hspace*{5mm}if PremiseMacro $\in s$ then  { \\
							\hspace*{6mm}$Premise \leftarrow words[ofstPremiseMacro, ofstCC]$ }\\
						  			}
				 } 
	  
	} else{ \\
		\hspace*{2mm}if ClaimMacro $\in s$ then { \\
				\hspace*{4mm}$Claim \leftarrow words[ofstClaimMacro, ofstPct]$ \\
		\hspace*{1mm} else { \\
					\hspace*{4mm}if PremiseMacro $\in s$ then  { \\
							\hspace*{5mm}$Premise \leftarrow words[ofstPremiseMacro, ofstPct]$ }\\
						}
					}
		}
}
}
\end{algorithm}

The top level approach for detecting arguments is formalised in Algorithm~\ref{alg:eval}.
First, the system analysis every documents $d$ from the corpus $\mathcal{C}$ of available medical documents (line 1). Each document $d$ is tokenised (line 3) and for each sentence $s$ the system verifies if a coordinating conjunction ($CC$) exists in the sentence (line 4). 
If such $CC$ is found, the algorithm searches an instance of the concept \textit{ClaimMacro}, as a possible indicator for an argument claim (line 5). 
The tool looks after the conjunction and verifies if the \textit{PremiseMacro} is present, then the \textit{Premise} is identified (lines 8 and 9). If the offset of the \textit{ClaimMacro} is higher than the CC offset than the system looks for the \textit{Claim} and \textit{Premise} inside the sentences (lines 11, 12 and 13). 

If within sentences does not contain a coordinating conjunction (line 14) then an instance of \textit{ClaimMacro} or \textit{PremiseMacro} is searched. Depending on its presence the tool determines whether that sentence can be associated to an annotation between \textit{Claim} or \textit{Premise} (lines 15, 16, 17, 18, 19). If the sentence does not contain a coordinating conjunction or \textit{ClaimMacro} or \textit{PremiseMacro} than the system analyses the next sentence in the set of \textit{Sentences}. 

The system uses macros, indicators to decide if a sentence or a part of this sentence is a potential candidate to \textit{Claim} or \textit{Premise}.
	 
The following templates for claim were used:
	 i) \textit{Ci}: claim indicator; %illustrated in Example~\ref{ex:claim}
	 ii) \textit{CbPe}: claim indicator plus people involved; %exemplified in example~\ref{ex:claimIndicPlusPeople}
	 iii) \textit{CbPebVc}: claim indicators followed by people involved and specific verbs to conclusion; %presented in example~\ref{ex:claimIndicPeopleVerbs}
	 iiii) \textit{CbVc}: people involved and than verbs for claim; %referenced in example~\ref{ex:claimIndicVerbs}
	v) \textit{ElOfCnbCw}: word or words expressing elements of cancer followed by a word that refers to cancer; after this macro can be present or not a verb that refers to claim; %this is noted in example~\ref{ex:claimIndicFactors}
	v) \textit{CbQ}: people can have quantifiers before them representing the number of them. %illustrated in example~\ref{ex:quantifiers}
We rely on the textual indicators, classified as follows :
$Ci \sqcup CibPe \sqcup CibPebVc \sqcup CibVc \sqcup ElOfCnbCw \sqcup CbQ \sqsubseteq ClaimIndicator
ClaimIndicator \sqsubseteq MacroIndicator$.

%\begin{example}[ClaimIdentifier] \textit{ClaimIndicator} is a concept presented above in the example~\ref{ex:claim}.
%\label{ex:claimIndic}
%\end{example}
The macro \textit{ClaimIndicator before People (MI\_CbPe)} %The following expressions are instances of the concept that 
contains \textit{ClaimIndicator} succeeded by people involved in the medical domain of breast cancer:

\begin{footnotesize}
\begin{schema}{MI: \text{ClaimIndicator before People}\doteqdot MI\_CibPe}
	ci:ClaimIndicator\\
	pe:Person\\
	(ci,pe):before\\
	(sentence,ci):hasToken\\
	(sentence,pe):hasToken
   \where
   Ex_1: \text{["we may [infer]$_{ci}$ that [woman]$_{pe}$"]$_{CibPe}$}\\
   Ex_2: \text{["this [bears out]$_{ci}$ the point that [doctors]$_{pe}$"]$_{CibPe}$}\\
   Ex_3: \text{["it [follows]$_{ci}$ that [patiences]$_{pe}$"]$_{CibPe}$}\\
   \end{schema}
\end{footnotesize} 

The sentence $Ex_1$ is formalised as an instance of MI\_CibPe in Example~\ref{ex:claimIndicPlusPeople}.  

\begin{example}
\textit{CibPe} macro contains individuals $ci$ of type \textit{ClaimIndicator} and $pe$ of type \textit{Person}. Within $sentence$ (lines \ref{eq:13}, \ref{eq:14}), $ci$ is located before $pe$ (line \ref{eq:12}). The text of both individuals is  presented via the role \textit{hasText} (line~\ref{eq:15}).
   
\begin{small}
\begin{align}   
   ci:ClaimIndicator, pe:Person \label{eq:11}\\
   (ci,Pe):before \label{eq:12}\\
   (sentence,ci):hasToken \label{eq:13}\\
   (sentence,pe):hasToken \label{eq:14}\\ 
   (ci,"infer"):hasText, (pe,"woman"):hasText \label{eq:16}
\end{align}
\end{small}
\label{ex:claimIndicPlusPeople}
\end{example}

The macro \textit{ClaimIndicator before People before Verb related to Claim (MI\_CibPebVc)} %The following expressions are instances of macro showed above 
is a generalisation of the macro \textit{ClaimIndicator before People (MI\_CibPe)} plus verbs related to conclusion (recall Example~\ref{ex:VerbClaim}): 

\begin{footnotesize}
\begin{schema}{MI: \text{ClaimIndicator before People before Verb}\doteqdot MI\_CibPebVc}
    ci:ClaimIndicator\\
   	pe:Person\\
	vc:VerbRelatedToClaim\\
    (ci,pe):before\\
    (pe,vc):before\\
	(sentence,ci):hasToken\\
	(sentence,pe):hasToken\\
	(sentence,vc):hasToken
   \where
   Ex_1: \text{["we can [conclude]$_{ci}$ that [doctors]$_{pe}$ [identified]$_{vc}$"]$_{CibPebVc}$}\\
   Ex_2: \text{["[so]$_{ci}$ the [key informants]$_{pe}$ [provides]$_{vc}$"]$_{CibPebVc}$}\\
   Ex_3: \text{["it [follows]$_{ci}$ that [people]$_{pe}$ [estimated]$_{vc}$"]$_{CibPebVc}$}\\
   \end{schema}
\end{footnotesize} 

%\label{ex:claimIndicPeopleVerbs}

The macro \textit{ClaimIndicator before Verb related to Claim (MI\_CbVc)} contains expressions that are instances of the concept \textit{ClaimIndicator} followed by verbs related to conclusion (recall Example~\ref{ex:VerbClaim}): 

\begin{footnotesize}
\begin{schema}{MI: \text{ClaimIndicator before Verb}\doteqdot MI\_CibVc}
  ci:ClaimIndicator\\
  vc:VerbRelatedToClaim\\
 (ci,vc):before\\
 (sentence,ci):hasToken\\
 (sentence,vc):hasToken
   \where
   Ex_1: \text{["[therefore]$_{ci}$ [exemplifies]$_{vc}$"]$_{CibVc}$}\\
   Ex_2: \text{["[so]$_{ci}$ [highlighted]$_{vc}$"]$_{CibVc}$}\\
   Ex_3: \text{["[thus]$_{ci}$ [accepted]$_{vc}$"]$_{CibVc}$}\\
   \end{schema}
\end{footnotesize} 

%\label{ex:claimIndicVerbs}
%\end{example}

The expression $Ex_1$ is formalised as an instance of  MI\_CbVc as Example~\ref{ex:claimIndicVerbs} illustrates. 

\begin{example}
\textit{CibVc} macro contains the individuals $ci$ of type \textit{ClaimIndicator} and $vc$ of type \textit{VerbRelatedToClaim}. Inside the $sentence$ (line \ref{eq:21}), $ci$ is located before $vc$ (line \ref{eq:19}). 
The text of both individuals is presented via the role \textit{hasText} (line \ref{eq:23}).
\begin{small}
\begin{align}
    ci:ClaimIndicator, vc:VerbRelatedToClaim \label{eq:18}\\
   (ci,vc):before \label{eq:19}\\
   (sentence,ci):hasToken, (sentence,vc):hasToken \label{eq:21}\\ 
   (ci,"therefore"):hasText, (vc,"exemplifies"):hasText \label{eq:23}
\end{align}
\end{small}
\label{ex:claimIndicVerbs}
\end{example}
The macro \textit{Elements of Cancer before Cancer related words (MI\_ElOfCnbCwVc)} contains expressions that are composed by one or more words that are instances of the concept \textit{ElementsOfCancer} plus a word from the cancer domain and optional can be succeeded by a verb related to claim (recall Example~\ref{ex:VerbClaim}): 

\begin{footnotesize}
\begin{schema}{MI: \text{Elements of Cancer before Cancer words}\doteqdot MI\_ElOfCnbCw}
  elOfCn:ElementsOfCancer\\
  cw:CancerWords\\
 (elOfCn,cw):before\\
 (sentence,elOfCn):hasToken\\
 (sentence,cw):hasToken
   \where
   Ex_1: \text{["the [risk]$_{elOfCn}$ of [breast cancer]$_{cw}$"]$_{ElOfCnbCwVc}$}\\
   Ex_2: \text{["these [factors]$_{elOfCn}$ of [cancer]$_{cw}$ [were equaled]$_{vc}$"]$_{ElOfCnbCwVc}$}\\
   \end{schema}
\end{footnotesize} 

%\label{ex:claimIndicFactors}

The macro \textit{Qualifiers before People (MI\_QbPebVc)} contains qualifiers before the instances of the concept \textit{PeopleInvolved}, meaning that more people are involved : 

\begin{footnotesize}
\begin{schema}{MI: \text{Qualifiers before People before Verb}\doteqdot MI\_QbPebVc}
   q:Qualifier\\
   pe:Person\\
   vc:VerbRelatedToClaim\\
  (q,pe):before\\
  (pe,vc):before\\
  (sentence,q):hasToken\\
  (sentence,pe):hasToken\\
  (sentence,vc):hasToken
   \where
   Ex_1: \text{["[many]$_{q}$ [woman]$_{pe}$ [provides]$_{vc}$"]$_{QbPebVc}$}\\
   Ex_2: \text{["[many]$_{q}$ [survivors]$_{pe}$ [accepted]$_{vc}$"]$_{QbPebVc}$}\\
   \end{schema}
\end{footnotesize}

%\label{ex:quantifiers}

The following templates for premises were searched:
	 i) \textit{PibPe}: premise indicators; %illustrated in Example~\ref{ex:premise}
	 ii) \textit{PibPebVp}: premise indicators plus people involved; %presented in example~\ref{ex:premiseIndicPlusPeople}
	 iii) \textit{PibPebVp}: premise indicators followed by people involved and specific verbs to premise; %presented in example~\ref{ex:premiseIndicPeopleVerbs}
	 iiii) \textit{PibVp}: people involved and than verbs for premise %referenced in example~\ref{ex:premiseIndicVerbs}
	v) \textit{ElOfCnbCw}: word or words expressing elements of cancer followed by a word that refers to cancer; after this macro can be present a verb that refers to claim or not; %noted in example~\ref{ex:premiseIndicFactors}
	 vi) \textit{DbVp}: words that express domains affected of breast cancer followed by verbs specific to premise; %presented in example \ref{ex:domainsAffected}	 
%\begin{example}[PremiseIdentifier]\textit{PremiseIndicator} is a concept presented above in the example~\ref{ex:premise}.
%\label{ex:premiseIndic}
%\end{example}
We rely on the textual indicators:
$Pi \cup PibPe \sqcup PibPebVp \sqcup PibVp \sqcup ElOfCnbCW \sqcup DbVp \sqsubseteq PremiseIndicator
PremiseIndicator \sqsubseteq MacroIndicator$.

The macro \textit{PremiseIndicator before People (MI\_PibPe)} contains expressions that are instances of the concept that contains \textit{PremiseIndicator} followed by people involved in the medical domain of breast cancer :

\begin{footnotesize}
\begin{schema}{MI: \text{PremiseIndicator before People}\doteqdot MI\_PibPe}
    pi:PremiseIndicator\\
	pe:Person\\
	(pi,pe):before\\
	(sentence,pi):hasToken\\
	(sentence,pe):hasToken
   \where
   Ex_1: \text{["[in view of the fact]$_{pi}$ that [woman]$_{pe}$"]$_{PibPe}$}\\
   Ex_2: \text{["[as shown] $_{pi}$ by [doctors]$_{pe}$"]$_{PibPe}$}\\
   Ex_3: \text{["[since]$_{pi}$ [patiences]$_{pe}$"]$_{PibPe}$}\\
   \end{schema}
\end{footnotesize}

%\label{ex:premiseIndicPlusPeople}

The macro \textit{PremiseIndicator before People before Verb related to premise (MI\_PibPebVp)} is a generalisation of the macro \textit{PremiseIndicator before People (MI\_PibPe)} plus verbs related to premise (recall Example~\ref{ex:VerbPremise}): 

\begin{footnotesize}
\begin{schema}{MI: \text{PremiseIndicator before People before Verb}\doteqdot MI\_PibPebVp}
   pi:PremiseIndicator\\
   pe:People\\
   vp:VerbRelatedToPremise\\
  (pi,pe):before, (pe,vp):before\\
  (sentence,pi):hasToken\\
  (sentence,vp):hasToken\\
  (sentence,pe):hasToken
   \where
   Ex_1: \text{["as [evidenced]$_{pi}$ by [people]$_{pe}$ [received]$_{vp}$"]$_{PibPebVp}$}\\
   Ex_2: \text{["[assuming]$_{pi}$ that [doctors]$_{pe}$ [observed]$_{vp}$"]$_{PibPebVp}$}\\
   Ex_3: \text{["[because]$_{pi}$ the [key informants]$_{pe}$ [were noted]$_{vp}$"]$_{PibPebVp}$}\\
   \end{schema}
\end{footnotesize} 
 
The expression $Ex_1$ is formalised as an instance of the MI\_PibPebVp macro indicator, as Example~\ref{ex:premiseIndicPeopleVerbs} illustrates. 
\begin{example}
\textit{PibPebVp} macro contains the individuals $pi$ of type \textit{PremiseIndicator}, $pe$ of type \textit{People} and $vp$ of type \textit{VerbRelatedToPremise}. 
$pi$ is located before $Pe$ and $pe$ is located before $vp$ (line \ref{eq:28}). 
%The text of the individuals are presented via the role \textit{hasText} (line~\ref{eq:34}).
\begin{small}
\begin{align}
  pi:PremiseIndicator, pe:People, vp:VerbRelatedToPremise \label{eq:26}\\
  (pi,pe):before, (pe,vp):before \label{eq:28}\\
  (sentence,pi):hasToken, (sentence,vp):hasToken \label{eq:30}\\
  (sentence,pe):hasToken \label{eq:31}\\
  (pi,"evidenced"):hasText, (pe,"people"):hasText\\
  (vp,"received"):hasText \label{eq:34}
\end{align}
\end{small}
\label{ex:premiseIndicPeopleVerbs}
\end{example}

The macro \textit{PremiseIndicator before Verb (MI\_PibVp)} contains expressions that are instances of the concept \textit{PremiseIndicator} succeeded by verbs related to premise: %(recall Example~\ref{ex:VerbPremise}): 
\begin{footnotesize}
\begin{schema}{MI: \text{PremiseIndicator before Verb}\doteqdot MI\_PibVp}
   pi:PremiseIndicator\\
   vp:VerbRelatedToPremise\\
  (pi,vp):before\\
  (sentence,pi):hasToken\\
  (sentence,vp):hasToken
   \where
   Ex_1: \text{["[since]$_{pi}$ [according]$_{vp}$"]$_{PibVp}$}\\
   Ex_2: \text{["[given]$_{pi}$ that [noted]$_{vp}$"]$_{PibVp}$}\\
   Ex_3: \text{["[seeing]$_{pi}$ that [served]$_{vp}$"]$_{PibVp}$}\\
   \end{schema}
\end{footnotesize}

%\label{ex:premiseIndicVerbs}

The macro \textit{Elements of Cancer before Cancer related words (MI\_ElOfCnbCw)} contains the expressions that are composed by one or more words that are instances of the concept \textit{ElementsOfCancer} plus a word from the cancer domain and optional can be succeeded by a verb related to premise (recall Example~\ref{ex:VerbPremise}): 

\begin{footnotesize}
\begin{schema}{MI: \text{Elements of Cancer before Cancer words}\doteqdot MI\_ElOfCnbCw}
  elOfCn:ElementsOfCancer\\
  cw:CancerWords\\
 (elOfCn,cw):before\\
 (sentence,elOfCn):hasToken\\
 (sentence,cw):hasToken
   \where
   Ex_1: \text{"[the [risk]$_{elOfCn}$ of [breast cancer]$_{cw}$ [was noted]$_{vp}$"]$_{ElOfCnbCw}$}\\
   \end{schema}
\end{footnotesize} 

%\label{ex:premiseIndicFactors}

The macro \textit{Domains affected before Verb (MI\_DbVp)} contains expressions that are composed by domains affected by breast cancer followed by a verb related to premise. % recalled in the example~\ref{ex:VerbPremise}:  

\begin{footnotesize}
\begin{schema}{MI: \text{Domains affected before Verb}\doteqdot MI\_DbVp}
  	d:Domains\\
  	vp:VerbRelatedToPremise\\
   (d,vp):before\\
   (sentence,d):hasToken\\
   (sentence,vp):hasToken
   \where
   Ex_1: \text{["[Family history]$_{d}$ [regarding]$_{vp}$"]$_{DbVp}$}\\ 
   Ex_2: \text{["[physical changes]$_{d}$ [resulting]$_{vp}$"]$_{DbVp}$}\\ 
   \end{schema}
\end{footnotesize}
   
The expression $Ex_1$ is formalised as an instance of  MI\_DbVp as example~\ref{ex:domainsAffected} illustrates.

\begin{example}
\textit{DbVp} macro contains the individuals $d$ of type \textit{Domains} and $vp$ of type \textit{VerbRelatedToPremise}. Inside the $sentence$ (line~\ref{eq:39}), $d$ is located before $vp$ (line \ref{eq:37}). 
The text of the individuals are presented via the role \textit{hasText} (line~\ref{eq:41}).
\begin{small}
\begin{align}
   d:Domains, vp:VerbRelatedToPremise, (d,vp):before \label{eq:37}\\
   (sentence,d):hasToken,(sentence,vp):hasToken \label{eq:39}\\
   (d,"Family history"):hasText, (vp,"regarding"):hasText \label{eq:41}
\end{align}
\end{small}
\label{ex:domainsAffected}
\end{example}

\section{Running experiments}
\label{sec:experiment}

%If the detection of the argumentative propositions of a text is possible, then it seems that the %classification of these propositions by their argumentative function should also be feasible.

\begin{figure}
\includegraphics[width=0.5\textwidth]{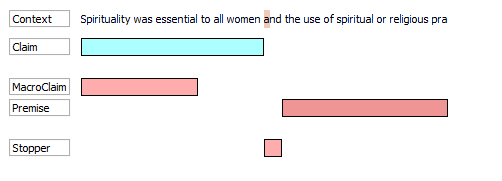}
\caption{Sample output from an ANNIE search}
\label{fig:sample}
\end{figure}

To identify text fragments that can be used to instantiate the argumentation schemes, we use ANNIE to investigate the entire corpus. Figure~\ref{fig:sample} shows a result of a search for \textit{Claim} and \textit{Premise} annotation, where the "Context" represents a sentences from a document. 
Based on the Jape rules, the system need to know if the coordinating conjunction and the macro of claim or premise is present inside the sentence.

\begin{figure*}
\begin{center}
\includegraphics[width=0.95\textwidth]{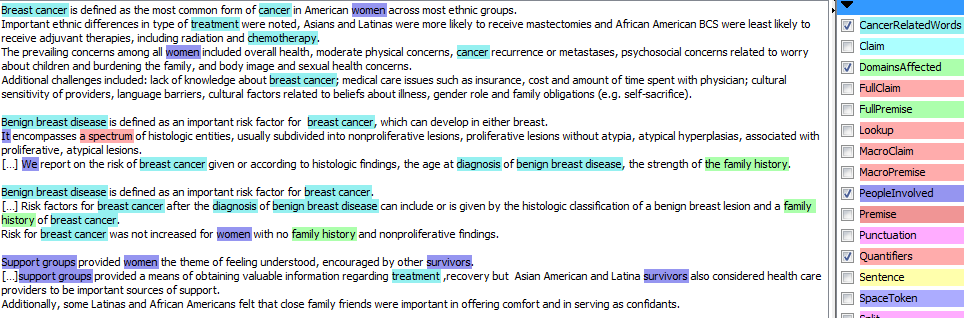} 
\end{center}
\caption{An annotated text from breast cancer articles.}
\label{fig:annotated}
\end{figure*}

We can also look at annotations inside a text. 
Figure~\ref{fig:annotated} shows one paragraph of a document, with a variety of annotation types, where are highlighted different annotation types; from this text the tool extracts word related to \textit{Cancer}, \textit{People involved}, \textit{Domains affected} by breast cancer and \textit{Qualifiers}.

\label{sec:related}

\label{sec:related}

The arguments were identified throw a corpus formed by six text documents related to breast cancer. In every text document are identified between five and ten arguments.

The quantitative evaluation is based on the measure of the \textit{Precision}, \textit{Recall} and \textit{F-Measure} metrics. The system was evaluated on a manually annotated corpus containing six documents with different breast cancer articles. The quantitative metrics were obtained with the \textit{Diff} plugin, integrated in GATE, applied to each document %\footnote{available at }. 
The percentage obtained for \textit{Claim}and for \textit{Premise} identification in Table~\ref{tab:claim}.
%In pattern recognition and information retrieval with binary classification, precision is the fraction %of retrieved instances that are relevant, while recall is the fraction of relevant instances that are %retrieved. 
The results obtained by the system are influenced by the performance of the correct identification of different parts of speech. Arguments are identified by the application according to lists of words and part of speech obtained by the MiniPar Parser included in GATE. If the annotations are not correctly identified this will limit the performance of the system.

\begin{table}
\begin{footnotesize}
\caption{Identifying claim of arguments (left) and premises (right).}
\label{tab:claim}
\begin{center}
\begin{tabular}{|l|l|l|l|}
\hline
{\it \#} & {\it Recall} & {\it Precision} & {\it F} \\
\hline
1 & 0.875 & 0.715 & 0.775\\ \hline 
2 & 0.815 & 0.665 & 0.725\\ \hline
3 & 0.75 & 0.675 & 0.71 \\ \hline
4 & 0.75 & 0.9 & 0.86 \\ \hline
5 & 0.9 & 0.65 & 0.78 \\ \hline
6 & 0.88 & 0.715 & 0.775 \\ \hline
\end{tabular}
%\end{center}
%\end{footnotesize}
%\end{table}
%\begin{table}
%\begin{footnotesize}
%\caption{Results from the premise of arguments inside text documents.}
%\label{tab:premise}
%\begin{center}
\begin{tabular}{|l|l|l|l|}
\hline
{\it \#} & {\it Recall} & {\it Precision} & {\it F} \\
\hline
1 & 0.69 & 0.75 & 0.65\\ \hline 
2 & 0.85 & 0.58 & 0.7\\ \hline
3 & 0.91 & 0.68 & 0.75 \\ \hline
4 & 0.75 & 0.9 & 0.86 \\ \hline
5 & 0.9 & 0.75 & 0.83 \\ \hline
6 & 0.96 & 0.76 & 0.83 \\ \hline
\end{tabular}
\end{center}
\end{footnotesize}
\end{table}

\section{Related work and discussion}
\label{sec:related}

There are several tools used for identifying arguments inside texts using natural language processing. %Also, this fields of identification have been studied in a wide variety of contexts. 
Rules have been extracted from scientific papers using SWRL in Controlled English (SR-CE) in FluentEditor~\cite{wroblewska2013semantic}. 
\cite{genetics2015} has proposed a specification of ten causal argumentation schemes used to detect arguments for scientific claims in genetics research journal articles. 
The specifications and some of the examples from which they were derived were used to create an initial draft of guidelines for annotation of a corpus. 
%The guidelines were evaluated in a pilot study that showed that several key schemes could be recognized by other %researchers based upon reading the guidelines.
Feng and Hirst~\cite{classify} have investigated argumentation scheme recognition using the Araucaria corpus, which contains annotated arguments from newspaper articles, parliamentary records, magazines, and on-line discussion boards (Reed et al. 2010). Taking premises and conclusion as given, Feng and Hirst addressed the problem of recognizing the name of the argumentation scheme for the five most frequently occurring schemes of Walton~\cite{Walton} in the corpus: Argument from example, Argument from cause to effect, Practical reasoning, Argument from Consequences, and Argument from Verbal Classification.

Other applications~\cite{consumersArg} have used annotations made by hand. 
There is no automatic detection of annotations, discourse indicators as well as user, domain, and sentiment terminology being identify manually. 
The difference between our system and this tool is based on this identification. Our application uses JAPE rules implemented in GATE for the identification of the claim and premise. %This is an advantage, but in the same time a disadvantage because the incorrect detection of a word that a rule relies can usually result in inaccurate.
Other researchers~\cite{argumentsWorbench2015} discuss the architecture and development of an Argument Workbench, which is a interactive, integrated, modular tool set to extract, reconstruct, and visualise arguments.
The Argument Workbench supports an argument engineer to reconstruct arguments from textual sources, using information processed at one stage as input to a subsequent stage of analysis, and then building an argument graph. The tool harvest and preprocess comments; highlight argument indicators, speech act and epistemic terminology; model topics; and identify domain terminology. The argument engineer analysis the output and then the input is extracted into the DebateGraph visualisation tool.

\section{Conclusion}
\label{sec:con}
Here we integrated ontologies and NLP for identifying arguments from breast cancer articles. %Even through we have focused only on detecting the arguments based on JAPE rules and lists of words, we were able to discover arguments with a high recall and precision. 
The contributions of this paper are: 
Firstly, we formalised an argumentation model in description logics. 
Hence, the arguments can be automatically classified, reasoning services of DL can be used on the model and the arguments can be retrieved by querying the ontology.
Secondly, we developed o tool able to perform argumentation mining. 
During mining, the tool uses concepts and roles within a breast cancer ontology. 
By changing the domain ontology, the tool can be applied to a different domain. 
%We are currently focusing on solving the problem of introducing other plugins, like WordNet~\cite{fellbaum1998wordnet} and 
%SimpleNLG~\cite{simpleNLG}, tools used for make more precisely the lists which the rules of detection are based. 

% For peer review papers, you can put extra information on the cover
% page as needed:
% \ifCLASSOPTIONpeerreview
% \begin{center} \bfseries EDICS Category: 3-BBND \end{center}
% \fi
%
% For peerreview papers, this IEEEtran command inserts a page break and
% creates the second title. It will be ignored for other modes.
\IEEEpeerreviewmaketitle

% use section* for acknowledgement
%\section*{Acknowledgements}

\bibliographystyle{IEEEtran}      % mathematics and physical sciences
\bibliography{bib}

% that's all folks
\end{document}